\documentclass[10pt]{article}

\usepackage[margin=1in]{geometry}
\usepackage{times}
\usepackage{microtype}
\usepackage{graphicx}
\usepackage{booktabs}
\usepackage{amsmath,amssymb}
\usepackage{enumitem}
\usepackage{algorithm}
\usepackage{algpseudocode}
\usepackage{float}
\usepackage{xcolor}
\usepackage{url}
\usepackage[hidelinks]{hyperref}

\title{\textbf{MDL-Calibrated Significance-Gain Pair Encoding:\\
Replication-Aware Automatic Stopping for Subword Tokenization}}

\author{Azam Nouri\\
\small Department of Science, Technology \& Mathematics,
Lincoln University, MO, USA}

\date{}

\begin{document}
\maketitle

\begin{abstract}

Byte-Pair Encoding (BPE) constructs subword vocabularies through greedy pair
merging, but conventional BPE requires the number of merges or target
vocabulary size to be specified externally. Significance-Gain Pair Encoding
(SG-BPE) replaces frequency-only selection with a statistical criterion that
measures how strongly an observed pair exceeds its expected co-occurrence
under an independence model \cite{nouri2026siggain}. Although this provides a
more statistically grounded ranking of candidate merges, statistical
association alone does not determine whether a newly created token remains
useful as tokenization proceeds. In particular, sparse pairs can exhibit
strong association because their expected counts become very small, even when
adding the corresponding token provides little practical benefit.

This paper introduces \emph{MDL-Calibrated Significance-Gain Pair Encoding}
(MDL-SG), a three-stage tokenization procedure that separates
\emph{discovery}, \emph{replication}, and \emph{utility}. Candidate pairs are
first ranked by the Significance-Gain criterion on a discovery partition.
The association is then tested for replication on a separate replication
partition using an exact one-sided hypergeometric test followed by
per-iteration Benjamini--Hochberg correction. Finally, a separate utility
partition evaluates whether the candidate produces a positive net Minimum
Description Length (MDL) gain after accounting for both sequence coding cost
and the cost of introducing a new merge rule. The merge process terminates
automatically when no statistically replicated candidate yields positive
held-out MDL gain.

Experiments on WikiText-103 demonstrate data-dependent stopping at 209, 433,
and 847 merges for tokenizer-training samples of 120K, 250K, and 500K
characters, respectively. At 500K characters, the method selects a stored
vocabulary of 1,017 tokens without prescribing that vocabulary size in
advance. In a compute-matched TinyGPT experiment using identical
2,024,448-parameter models and exactly 500 optimizer updates per language
model, MDL-SG achieves validation and test bits-per-character (BPC) values of
3.2612 and 3.2436, respectively. The corresponding test BPC values are 3.2894
for SG-BPE and 3.3493 for frequency BPE, giving MDL-SG improvements of
1.39\% and 3.16\%, respectively. Interestingly, frequency BPE achieves
stronger raw compression in terms of tokens per character, whereas MDL-SG
achieves better predictive efficiency, indicating that compression-oriented
merge selection and language-model utility need not coincide.

\end{abstract}


\section{Introduction}

Subword tokenization determines the discrete representation through which a
language model observes text. Its design influences sequence length,
vocabulary size, embedding and output-layer dimensions, computational
requirements, and the statistical regularities presented during model
training. Byte-Pair Encoding (BPE), originally introduced as a data-compression
procedure \cite{gage1994new} and later adapted to neural natural language
processing \cite{sennrich2016neural}, remains one of the most widely used
approaches for constructing subword vocabularies. Related tokenization
frameworks, including SentencePiece, have further enabled language-independent
and implementation-independent subword processing
\cite{kudo2018sentencepiece}.

Standard BPE repeatedly identifies the most frequent adjacent token pair and
replaces that pair with a new token. If $c_{xy}$ denotes the frequency of pair
$(x,y)$, the classical selection rule is

\begin{equation}
    (x^\ast,y^\ast)
    =
    \arg\max_{(x,y)} c_{xy}.
\end{equation}

This rule is computationally simple and naturally favors sequence compression.
However, high pair frequency does not necessarily imply strong pair-specific
dependence. Two individually common symbols can form a frequent pair largely
because both have high marginal frequencies. Moreover, conventional BPE does
not determine its own stopping point: a target vocabulary size or number of
merge operations is ordinarily chosen before tokenizer training.

The vocabulary-size decision is consequential. A small vocabulary produces
longer token sequences, whereas continued merging creates increasingly
specialized tokens whose statistical support may become weak. The appropriate
stopping point can depend on corpus size, domain, and the downstream objective.
Therefore, prescribing a single merge count independently of the observed
statistical structure leaves an important component of tokenizer construction
outside the learning procedure.

Statistically motivated alternatives have attempted to move beyond
frequency-only pair selection. Vilar and Federico proposed a statistical
extension of BPE in which candidate merges are assessed using statistical
association rather than frequency alone \cite{vilar2021statistical}. More
recently, Significance-Gain Pair Encoding (SG-BPE) introduced a merge score
based on the excess of observed adjacency over an independence-based
expectation while retaining an explicit support-sensitive gain term
\cite{nouri2026siggain}. For an observed pair count $c_{xy}$, marginal counts
$c_x$ and $c_y$, and $N$ adjacent positions, the expected count under the
independence model is

\begin{equation}
    E[c_{xy}]
    =
    \frac{c_xc_y}{N}.
    \label{eq:expected}
\end{equation}

The standardized excess association is then

\begin{equation}
    z(x,y)
    =
    \frac{
        c_{xy}-E[c_{xy}]
    }{
        \sqrt{E[c_{xy}]+\varepsilon}
    },
    \label{eq:zscore}
\end{equation}

where $\varepsilon>0$ is a fixed numerical stabilization constant used to
avoid division by zero when the expected count is extremely small.
For reproducibility, the implementation uses a single fixed value of
$\varepsilon$ throughout tokenizer training.\footnote{where $\varepsilon=...$ is a fixed numerical stabilization constant used
to avoid division by zero when the expected count is extremely small.}

The Significance-Gain score used in this work is

\begin{equation}
    S_{\mathrm{SG}}(x,y)
    =
    c_{xy}\,
    z(x,y)\,
    c_{xy}^{\alpha},
    \qquad
    \alpha=0.25.
    \label{eq:sgscore}
\end{equation}

The resulting ranking favors pairs that combine empirical support with
unexpectedly strong adjacency. In prior experiments, SG-BPE improved
character-normalized language-model performance relative to frequency BPE
across the principal operating point and most comparisons at approximately
matched compression levels
\cite{nouri2026siggain}. The present work addresses a separate limitation:
a statistically meaningful ranking criterion does not by itself determine
when the tokenizer should stop merging.

This limitation becomes particularly important during later merge iterations.
As previously created tokens become more specialized, candidate pairs may
occur only a modest number of times. Their expected counts under the
independence model can nevertheless become extremely small. Consequently, a
pair may exhibit a large standardized association even though introducing it
as an additional vocabulary item provides little useful reduction in
representation cost. In other words,

\begin{equation}
    \text{statistical association}
    \;\not\Rightarrow\;
    \text{useful new token}.
\end{equation}

A stopping rule based only on statistical significance can therefore continue
accepting increasingly sparse structures. Conversely, using a manually tuned
effect-size or support threshold \emph{as the stopping mechanism} would largely
replace the target-vocabulary hyperparameter with a different externally
chosen stopping threshold. The minimum pair-count requirement used later in
this work instead serves only as a fixed candidate-screening criterion and
does not determine when tokenizer training terminates.

The approach developed here separates the merge decision into three distinct
questions. First, \emph{discovery} asks whether a candidate is strongly
supported by the statistical structure of one portion of the corpus. Second,
\emph{replication} asks whether the same pairwise association persists in a
separate, disjoint portion of the tokenizer-training corpus. Third,
\emph{utility} asks whether introducing the replicated pair actually reduces
held-out description length after accounting for the complexity of the
additional token.

The tokenizer-training corpus is therefore divided into three disjoint
partitions,

\begin{equation}
    \mathcal{D}
    =
    \mathcal{D}_{\mathrm{disc}}
    \cup
    \mathcal{D}_{\mathrm{rep}}
    \cup
    \mathcal{D}_{\mathrm{util}},
\end{equation}

using proportions 70\%, 15\%, and 15\%, respectively. Significance-Gain is
computed only on the discovery partition. Candidate pairs are then tested for
replication on the separate replication partition using an exact one-sided
hypergeometric test. Because thousands of candidate pairs may be examined
within an iteration, their $p$-values are corrected using the
Benjamini--Hochberg procedure at $q=0.05$
\cite{benjamini1995controlling}. The correction is applied separately within
each merge iteration; no global false-discovery-rate guarantee over the full
adaptive merge trajectory is claimed.

Statistically replicated pairs then enter a held-out MDL utility test. Let the
utility sequence contain $T$ tokens with counts $n_1,\ldots,n_V$. Its empirical
data code length is

\begin{equation}
    L_{\mathrm{data}}
    =
    T\log_2 T
    -
    \sum_{j=1}^{V}
    n_j\log_2 n_j.
    \label{eq:data-mdl}
\end{equation}

To account for the complexity associated with growth of the active token
inventory, a BIC-style penalty is included \cite{schwarz1978estimating}:

\begin{equation}
    L_{\mathrm{BIC}}
    =
    \frac{V-1}{2}\log_2 T.
    \label{eq:bic}
\end{equation}

The total held-out description-length objective is therefore

\begin{equation}
    L_{\mathrm{total}}
    =
    L_{\mathrm{data}}
    +
    L_{\mathrm{BIC}}.
\end{equation}

This formulation follows the general Minimum Description Length principle that
a useful representation should balance goodness of fit against the complexity
required to describe the model or representation
\cite{rissanen1978modeling,grunwald2007mdl}.

Creating a merge also requires storing the identities of its two constituent
tokens. If $V_s$ denotes the current stored vocabulary size, the merge-rule
cost is

\begin{equation}
    C_{\mathrm{rule}}
    =
    2
    \left\lceil
    \log_2 V_s
    \right\rceil.
    \label{eq:rulecost}
\end{equation}

For a candidate merge, the net held-out MDL gain is defined as

\begin{equation}
    G_{\mathrm{MDL}}
    =
    L_{\mathrm{total}}^{\mathrm{before}}
    -
    L_{\mathrm{total}}^{\mathrm{after}}
    -
    C_{\mathrm{rule}}.
    \label{eq:netgain}
\end{equation}

A merge is eligible only when

\begin{equation}
    G_{\mathrm{MDL}} > 0.
\end{equation}

Thus, Significance-Gain determines the ranking of discovered candidates,
replication on a disjoint corpus partition determines statistical eligibility,
and held-out MDL determines whether the additional token provides sufficient
coding benefit to justify its added representation cost. The procedure
terminates when replicated candidates remain but none yields positive net
MDL gain.

\paragraph{Contributions.}
The principal contributions of this work are:

\begin{enumerate}[leftmargin=*,itemsep=2pt,topsep=3pt]

    \item A three-stage subword construction procedure that explicitly
    separates statistical discovery, replication on a disjoint data
    partition, and held-out utility evaluation.

    \item An exact replication test for candidate merges using directional
    marginal counts and one-sided hypergeometric probabilities, followed by
    per-iteration Benjamini--Hochberg correction.

    \item A held-out MDL acceptance criterion in which the zero threshold has
    a direct coding interpretation rather than being an empirically tuned
    stopping hyperparameter.

    \item An automatic stopping mechanism that selects different vocabulary
    sizes as the amount of tokenizer-training data increases. On 120K, 250K,
    and 500K WikiText-103 character samples, the procedure stops at 209, 433,
    and 847 merges, respectively, with the best remaining replicated
    candidates having net MDL gains of $-0.4223$, $-0.1507$, and $-0.0474$
    bits.

    \item A compute-matched language-model comparison showing that, at the
    500K-character operating point, MDL-SG achieves lower BPC than both
    frequency BPE and SG-BPE using the same model architecture and exactly
    500 optimizer updates per language model.

\end{enumerate}

The experimental results further reveal that the tokenizer achieving the
strongest raw compression is not necessarily the tokenizer producing the best
character-normalized language-model performance. At the matched 847-merge
budget, frequency BPE achieves a lower test tokens-per-character ratio than
MDL-SG, whereas MDL-SG obtains the lowest test BPC. This distinction
motivates evaluating subword vocabularies not only by sequence compression,
but also by the predictive usefulness of the representations they induce.

\section{Methodology}
\label{sec:method}

MDL-SG extends Significance-Gain Pair Encoding by separating each merge
decision into three stages: statistical discovery, replication on a disjoint
corpus partition, and held-out utility evaluation. The tokenizer begins from
a character-level vocabulary. After each accepted merge, the same merge is
applied to all three partitions so that their token inventories remain
synchronized throughout the procedure.

\subsection{Data Partitioning}
\label{subsec:partition}

Let the normalized tokenizer-training corpus be denoted by
$\mathcal{D}$. It is partitioned into three disjoint subsets,

\begin{equation}
    \mathcal{D}
    =
    \mathcal{D}_{\mathrm{disc}}
    \cup
    \mathcal{D}_{\mathrm{rep}}
    \cup
    \mathcal{D}_{\mathrm{util}},
\end{equation}

with proportions

\begin{equation}
    0.70,\qquad 0.15,\qquad 0.15,
\end{equation}

for discovery, replication, and utility, respectively. The discovery
partition determines the ranking of candidate merges. The replication
partition provides a separate statistical check, and the utility partition is
reserved for determining whether a statistically replicated candidate
provides positive held-out coding benefit.

This separation is important because using the same observations both to
discover a pair and to evaluate whether it replicates can exaggerate the
apparent strength of the association. Likewise, using the discovery data to
determine whether a token improves description length would allow the merge
criterion to optimize and evaluate itself on the same sample. Using disjoint
partitions therefore separates candidate ranking, replication testing, and
utility evaluation within the available tokenizer-training corpus.

\subsection{Stage I: Significance-Gain Discovery}
\label{subsec:discovery}

At iteration $t$, let the current discovery sequence be

\begin{equation}
    \mathbf{s}^{(t)}
    =
    (s_1,s_2,\ldots,s_T).
\end{equation}

For every adjacent pair $(x,y)$ satisfying the minimum candidate-support
requirement

\begin{equation}
    c_{xy} \geq c_{\min},
    \qquad c_{\min}=5,
\end{equation}

the pair count $c_{xy}$ and marginal token counts $c_x$ and $c_y$ are
computed. The value $c_{\min}=5$ serves only as a fixed candidate-screening
criterion; it is not used as the stopping rule for tokenizer training.

With $N=T-1$ available adjacent positions, the expected pair count under the
independence model is

\begin{equation}
    E[c_{xy}]
    =
    \frac{c_x c_y}{N}.
\end{equation}

The standardized excess association is

\begin{equation}
    z(x,y)
    =
    \frac{
        c_{xy}-E[c_{xy}]
    }{
        \sqrt{E[c_{xy}]+\varepsilon}
    },
\end{equation}

where $\varepsilon$ is the same fixed numerical stabilization constant defined
above.

Candidate pairs are ranked using

\begin{equation}
    S_{\mathrm{SG}}(x,y)
    =
    c_{xy}\,
    z(x,y)\,
    c_{xy}^{\alpha},
    \qquad
    \alpha=0.25.
    \label{eq:method-sg}
\end{equation}

The factor $c_{xy}^{\alpha}$ retains a controlled preference for empirically
supported pairs while allowing the standardized association term to
distinguish strongly cohesive pairs from pairs that are frequent primarily
because their constituent tokens are individually common.

Importantly, this stage determines only the \emph{ranking} of candidate
merges. A high Significance-Gain score alone is not sufficient for a merge
to be accepted.

\subsection{Stage II: Statistical Replication on a Disjoint Partition}
\label{subsec:replication}

Each discovery candidate is next evaluated on the separate replication
partition $\mathcal{D}_{\mathrm{rep}}$. Let $N_r$ denote the number of
adjacent positions in the current replication sequence. For a candidate
$(x,y)$, define

\begin{itemize}[leftmargin=*,itemsep=1pt,topsep=2pt]
    \item $k$ as the observed number of occurrences of pair $(x,y)$ in the
    replication partition;

    \item $n_x$ as the number of replication positions whose predecessor
    token is $x$;

    \item $K_y$ as the number of replication positions whose successor
    token is $y$.
\end{itemize}

Conditional on these directional marginals, the null distribution for the
pair count is

\begin{equation}
    X
    \sim
    \operatorname{Hypergeom}
    \left(
        N_r,\,
        K_y,\,
        n_x
    \right).
    \label{eq:hypergeom}
\end{equation}

The corresponding one-sided probability of observing at least $k$
occurrences under this null model is

\begin{equation}
    p_{xy}
    =
    \Pr(X\geq k)
    =
    \sum_{j=k}^{\min(n_x,K_y)}
    \frac{
        \binom{K_y}{j}
        \binom{N_r-K_y}{n_x-j}
    }{
        \binom{N_r}{n_x}
    }.
    \label{eq:rep-pvalue}
\end{equation}

Because many candidate pairs are tested at each merge iteration, the resulting
$p$-values are evaluated using the Benjamini--Hochberg (BH) multiple-testing
procedure \cite{benjamini1995controlling}. If $m$ hypotheses are tested and

\begin{equation}
    p_{(1)}
    \leq
    p_{(2)}
    \leq
    \cdots
    \leq
    p_{(m)}
\end{equation}

are the ordered $p$-values, let

\begin{equation}
    i^\ast
    =
    \max
    \left\{
        i:
        p_{(i)}
        \leq
        \frac{i}{m}q
    \right\},
    \qquad
    q=0.05.
    \label{eq:bh}
\end{equation}

When such an index exists, the hypotheses corresponding to
$p_{(1)},\ldots,p_{(i^\ast)}$ constitute the BH rejection set. Only candidate
pairs in this rejection set proceed to the utility stage. If no such index
exists, no candidate passes the replication stage in that iteration.

The BH procedure is applied separately within each merge iteration. Because
the candidate set changes adaptively after every accepted merge, no global
false-discovery-rate guarantee over the complete tokenizer-training trajectory
is claimed. Throughout this paper, ``replicated'' therefore refers to a
candidate that passes the specified one-sided replication test together with
the per-iteration BH procedure at $q=0.05$.

\subsection{Stage III: Held-Out MDL Utility}
\label{subsec:mdl}

Passing the replication stage indicates that the observed pair association
persists in a separate, disjoint portion of the tokenizer-training corpus.
However, statistical replication alone does not establish that creating a new
token provides sufficient representational benefit to justify its added
complexity. The third stage therefore evaluates each replicated candidate on
the held-out utility partition $\mathcal{D}_{\mathrm{util}}$.

Let the current utility sequence contain $T$ tokens and $V$ active token
types with counts $n_1,\ldots,n_V$. The empirical sequence code length is

\begin{equation}
    L_{\mathrm{data}}
    =
    T\log_2 T
    -
    \sum_{j=1}^{V}
    n_j\log_2 n_j.
    \label{eq:mdl-data}
\end{equation}

To account for growth of the active token inventory, the objective includes a
BIC-style complexity penalty \cite{schwarz1978estimating},

\begin{equation}
    L_{\mathrm{BIC}}
    =
    \frac{V-1}{2}\log_2 T.
    \label{eq:mdl-bic}
\end{equation}

The resulting held-out description-length objective is

\begin{equation}
    L_{\mathrm{total}}
    =
    L_{\mathrm{data}}
    +
    L_{\mathrm{BIC}}.
    \label{eq:mdl-total}
\end{equation}

If the current stored vocabulary contains $V_s$ entries, introducing the
merge $(x,y)\rightarrow z$ also incurs the dictionary cost

\begin{equation}
    C_{\mathrm{rule}}
    =
    2
    \left\lceil
        \log_2 V_s
    \right\rceil,
    \label{eq:mdl-rule}
\end{equation}

corresponding to two fixed-width references to existing stored tokens. The
candidate's net held-out MDL gain is then defined as

\begin{equation}
    G_{\mathrm{MDL}}(x,y)
    =
    L_{\mathrm{total}}^{\mathrm{before}}
    -
    L_{\mathrm{total}}^{\mathrm{after}}
    -
    C_{\mathrm{rule}}.
    \label{eq:mdl-final-gain}
\end{equation}

A candidate is eligible for merging only if

\begin{equation}
    G_{\mathrm{MDL}}(x,y)>0.
    \label{eq:positive-gain}
\end{equation}

The zero threshold is not an additional empirically tuned stopping
hyperparameter. It follows directly from the description-length objective:
a positive value means that the reduction in held-out description length
exceeds the cost assigned to introducing the new merge rule, whereas a
non-positive value means that the proposed token does not recover its added
representational cost.

Replicated candidates retain their Significance-Gain ordering. Beginning with
the highest-ranked replicated pair, the algorithm evaluates held-out MDL gain
until it encounters the first candidate satisfying
Eq.~\eqref{eq:positive-gain}. That candidate is accepted, and the same merge
is applied to the discovery, replication, and utility sequences. If no
replicated candidate has positive net MDL gain, tokenizer training terminates
automatically.

\subsection{Automatic Stopping Criterion}
\label{subsec:stopping}

The stopping rule follows directly from the held-out MDL objective. At
iteration $t$, let $\mathcal{R}^{(t)}$ denote the set of candidates that pass
the replication test and the per-iteration Benjamini--Hochberg procedure.
For candidates in $\mathcal{R}^{(t)}$, net held-out MDL gain is evaluated in
decreasing Significance-Gain rank order.

Tokenizer training continues if

\begin{equation}
    \mathcal{R}^{(t)} \neq \varnothing
    \quad\text{and}\quad
    \exists\, r\in\mathcal{R}^{(t)}
    \text{ such that }
    G_{\mathrm{MDL}}(r)>0.
    \label{eq:continue-condition}
\end{equation}

Conversely, the procedure terminates if either no candidate passes the
replication stage,

\begin{equation}
    \mathcal{R}^{(t)}=\varnothing,
\end{equation}

or replicated candidates remain but none provides positive net held-out MDL
gain,

\begin{equation}
    \max_{r\in\mathcal{R}^{(t)}}
    G_{\mathrm{MDL}}(r)
    \leq 0,
    \qquad
    \mathcal{R}^{(t)}\neq\varnothing.
    \label{eq:stopping-condition}
\end{equation}

This criterion differs fundamentally from a fixed merge budget. The tokenizer
is not required to produce a predetermined number of merges or vocabulary
items. Instead, the final vocabulary size emerges from the interaction among
the available corpus evidence, the replication criterion, and held-out coding
utility.

The implementation also includes maximum-vocabulary and maximum-merge limits
as computational safety caps. These limits are not part of the scientific
stopping criterion and were not reached in the reported 120K-, 250K-, or
500K-character experiments. In all three experiments, training terminated
because candidates continued to pass the replication stage but none had
positive net held-out MDL gain.

Algorithm~\ref{alg:mdlsg} summarizes the complete procedure.

\begin{algorithm}[H]
\caption{MDL-Calibrated Significance-Gain Pair Encoding}
\label{alg:mdlsg}
\begin{algorithmic}[1]

\Require Normalized corpus $\mathcal{D}$;
minimum candidate count $c_{\min}$;
SG exponent $\alpha$;
BH level $q$;
stabilization constant $\varepsilon$

\State Split $\mathcal{D}$ into disjoint partitions
$\mathcal{D}_{\mathrm{disc}}$,
$\mathcal{D}_{\mathrm{rep}}$, and
$\mathcal{D}_{\mathrm{util}}$

\State Initialize all three partitions using the same character vocabulary

\While{\textbf{true}}

    \State Count adjacent pairs in $\mathcal{D}_{\mathrm{disc}}$

    \State Remove candidates satisfying $c_{xy}<c_{\min}$

    \For{each remaining pair $(x,y)$}

        \State Compute
        $
        E[c_{xy}]
        =
        c_xc_y/N
        $

        \State Compute
        $
        z(x,y)
        =
        \dfrac{c_{xy}-E[c_{xy}]}
        {\sqrt{E[c_{xy}]+\varepsilon}}
        $

        \State Compute
        $
        S_{\mathrm{SG}}(x,y)
        =
        c_{xy}\,z(x,y)\,c_{xy}^{\alpha}
        $

    \EndFor

    \State Rank candidates by decreasing $S_{\mathrm{SG}}$

    \For{each discovery candidate $(x,y)$}

        \State Compute the exact one-sided hypergeometric
        $p$-value on $\mathcal{D}_{\mathrm{rep}}$

    \EndFor

    \State Apply the Benjamini--Hochberg procedure at level $q$

    \State Let $\mathcal{R}$ be the candidates that pass the
    replication stage

    \If{$\mathcal{R}=\varnothing$}
        \State \textbf{break}
    \EndIf

    \State $accepted \gets \mathrm{false}$

    \For{candidate $r$ in decreasing SG rank order within $\mathcal{R}$}

        \State Compute $G_{\mathrm{MDL}}(r)$ on
        $\mathcal{D}_{\mathrm{util}}$

        \If{$G_{\mathrm{MDL}}(r)>0$}

            \State Create a new token corresponding to candidate $r$

            \State Apply the same merge to
            $\mathcal{D}_{\mathrm{disc}}$,
            $\mathcal{D}_{\mathrm{rep}}$, and
            $\mathcal{D}_{\mathrm{util}}$

            \State $accepted \gets \mathrm{true}$

            \State \textbf{break}

        \EndIf

    \EndFor

    \If{$accepted=\mathrm{false}$}
        \State \textbf{break}
    \EndIf

\EndWhile

\State \Return learned merge sequence and resulting vocabulary

\end{algorithmic}
\end{algorithm}

\section{Experimental Setup}
\label{sec:experiments}

\subsection{Dataset and Preprocessing}
\label{subsec:data}

Experiments use the raw WikiText-103 corpus
\cite{merity2017pointer}. Tokenizer-training samples are drawn from the
training split, while language-model validation and test evaluations use the
separate WikiText-103 validation and test splits. Text is normalized by
collapsing repeated spaces and tabs and by limiting runs of three or more
newline characters to two newlines.

Three tokenizer-training sample sizes are considered:

\begin{equation}
    120{,}000,\qquad
    250{,}000,\qquad
    500{,}000
\end{equation}

characters. For each sample size, MDL-SG uses the same partition proportions:

\begin{equation}
    70\% \text{ discovery},
    \qquad
    15\% \text{ replication},
    \qquad
    15\% \text{ utility}.
\end{equation}

Thus, the 500K-character experiment contains 350K discovery characters,
75K replication characters, and 75K utility characters. The principal
tokenizer parameters are held fixed across corpus sizes:

\begin{equation}
    c_{\min}=5,
    \qquad
    \alpha=0.25,
    \qquad
    q=0.05.
\end{equation}

The same fixed numerical stabilization constant $\varepsilon$ defined earlier
is also used in every experiment. The maximum-vocabulary and maximum-merge
settings serve only as computational safety limits and are set above the
observed stopping points; neither limit is reached in the experiments
reported below.

\subsection{Automatic-Stopping Experiments}
\label{subsec:auto-stop}

The first experiment examines whether the same MDL-SG procedure selects
different stopping points as the amount of tokenizer-training data changes.
No target vocabulary size or merge count is supplied as the scientific
stopping rule.

Table~\ref{tab:scaling} summarizes the resulting numbers of accepted merges,
active vocabulary sizes, stored vocabulary sizes, and the largest MDL gain
among the replicated candidates rejected at the stopping boundary.

\begin{table}[H]
\centering
\caption{Automatic stopping behavior as tokenizer-training corpus size
increases. The final column reports the largest net MDL gain among replicated
candidates remaining at the stopping boundary.}
\label{tab:scaling}
\begin{tabular}{rrrrr}
\toprule
Corpus &
Merges &
Active vocab. &
Stored vocab. &
Best rejected gain (bits) \\
\midrule
120K & 209 & 327 & 348   & $-0.4223$ \\
250K & 433 & 565 & 583   & $-0.1507$ \\
500K & 847 & 976 & 1,017 & $-0.0474$ \\
\bottomrule
\end{tabular}
\end{table}

All three runs terminate under the
\texttt{no\_positive\_mdl\_gain} condition: candidates remain after the
replication stage, but none has positive net held-out MDL gain. The number of
accepted merges increases from 209 to 433 to 847 as the tokenizer-training
sample grows from 120K to 250K to 500K characters. Across these three tested
corpus sizes, this behavior is consistent with the intended data-dependent
stopping mechanism rather than a fixed externally imposed vocabulary size.

The stopping boundary was verified after the final accepted merge in each
experiment. Specifically, the complete set of discovery candidates was
recomputed, the replication procedure was applied again, and the held-out MDL
gain of every remaining replicated candidate was evaluated.

The largest remaining gains for the 120K-, 250K-, and 500K-character
experiments were

\begin{equation}
    -0.4223,\qquad
    -0.1507,\qquad
    -0.0474
    \quad \text{bits},
\end{equation}

respectively. Each value is non-positive, confirming that the corresponding
run satisfies the MDL stopping criterion. Importantly, stopping was not caused
by an absence of candidates passing the replication stage. Instead,
replicated candidates remained, but none provided sufficient held-out
description-length improvement to offset its assigned representation cost.

At the 500K-character stopping point, for example, 6,564 discovery candidates
remained, of which 1,890 passed the replication stage. Nevertheless, the
largest net MDL gain among those candidates was

\begin{equation}
    G_{\mathrm{MDL}}^{\max}
    =
    -0.0474\ \text{bits}.
\end{equation}

Therefore, no additional merge was accepted. This result illustrates the
distinction between passing the statistical replication criterion and
providing positive token utility under the held-out MDL objective.

\subsection{Baseline Tokenizers}
\label{subsec:baselines}

At the 500K-character operating point, MDL-SG automatically selects
847 merges and a stored vocabulary of 1,017 tokens. To isolate differences arising from merge selection, we compare MDL-SG
with two baseline tokenizers constructed using the same 847-merge budget:

\begin{enumerate}[leftmargin=*,itemsep=2pt,topsep=3pt]

    \item \textbf{Frequency BPE}, which repeatedly merges the
    highest-frequency eligible adjacent pair;

    \item \textbf{SG-BPE}, which ranks eligible pairs using the
    Significance-Gain criterion in Eq.~\eqref{eq:method-sg};

    \item \textbf{MDL-SG}, for which 847 merges are not imposed externally
    but arise from the proposed stopping procedure.

\end{enumerate}

Thus, the three methods are compared at the same number of accepted merges
and the same stored-vocabulary size. All three begin from the same
170-character base vocabulary and contain 1,017 stored tokens after
847 merges.

Their final \emph{active} vocabularies differ slightly because some tokens
created earlier in the merge sequence no longer occur in the final encoded
training sequence. The active vocabulary sizes are 1,002 for frequency BPE,
981 for SG-BPE, and 976 for MDL-SG.

For validation and test text, characters not observed in the 500K-character
tokenizer-training sample are mapped to a common \texttt{<UNK>} symbol.
Such cases are rare: 22 of 200,000 validation characters and 51 of 200,000
test characters. The same unknown-character handling is applied to all three
methods.

\subsection{Tokenizer-Only Evaluation}
\label{subsec:tokenizer-eval}

Before evaluating downstream language modeling, the three tokenizers are
compared directly in terms of the number of tokens required to represent the
same held-out text. Tokens per character (TPC) is defined as

\begin{equation}
    \mathrm{TPC}
    =
    \frac{
        \text{number of encoded tokens}
    }{
        \text{number of original characters}
    }.
    \label{eq:tpc}
\end{equation}

Lower TPC indicates stronger sequence compression. Table~\ref{tab:tpc}
reports TPC on 200K-character validation and test samples.

\begin{table}[H]
\centering
\caption{Tokenizer-only comparison at the matched 847-merge and
1,017-token stored-vocabulary operating point. Lower TPC indicates stronger
sequence compression.}
\label{tab:tpc}
\begin{tabular}{lrrrr}
\toprule
Method &
Val. tokens &
Val. TPC &
Test tokens &
Test TPC \\
\midrule
Frequency BPE & 76,153 & \textbf{0.380765} &
                75,592 & \textbf{0.377960} \\
SG-BPE        & 79,153 & 0.395765 &
                77,779 & 0.388895 \\
MDL-SG        & 80,042 & 0.400210 &
                79,002 & 0.395010 \\
\bottomrule
\end{tabular}
\end{table}

Frequency BPE produces the shortest token sequences on both validation and
test data. SG-BPE requires more tokens, and MDL-SG produces the longest
sequences among the three methods at this matched operating point.
Consequently, MDL-SG does not improve raw sequence compression relative to
the two baselines at the same stored-vocabulary size.

This distinction is important for interpreting the downstream language-model
results. MDL-SG does not directly optimize TPC. Instead, its utility stage
tests whether a statistically replicated merge reduces held-out description
length sufficiently to offset the assigned vocabulary-complexity and
merge-rule costs. A resulting vocabulary may therefore preserve somewhat
finer segmentation while still yielding a representation with different
predictive properties for next-token modeling.

\subsection{Language-Model Evaluation}
\label{subsec:lm-setup}

The downstream comparison uses the same small causal Transformer architecture
for all three tokenizers. Each language model contains four Transformer
blocks, four attention heads, a hidden dimension of 192, a context length of
256 tokens, and dropout probability 0.1. Token embeddings and the output
projection are weight-tied.

Because all three tokenizers have the same stored-vocabulary size and use one
additional \texttt{<UNK>} symbol during language-model evaluation, each model
has a vocabulary size of 1,018 and exactly 2,024,448 trainable parameters.

The three tokenized versions of the same 500K-character training sample have
different sequence lengths:

\begin{equation}
\begin{aligned}
    T_{\mathrm{Freq}} &= 179{,}765, \\
    T_{\mathrm{SG}}   &= 185{,}053, \\
    T_{\mathrm{MDL}}  &= 194{,}416.
\end{aligned}
\end{equation}

An epoch-matched comparison would therefore expose the three language models
to different numbers of optimization updates. To control the computational
training budget more directly, the primary downstream comparison instead
uses the same fixed number of optimizer updates for each tokenizer-specific
language model.

Each model is trained with AdamW using a learning rate of
$3\times10^{-4}$, weight decay 0.01, batch size 64, and randomly sampled
256-token training windows. The random seed is fixed at 42 for all three
models. Each model is trained for exactly 500 optimizer updates. Therefore,
each training run processes

\begin{equation}
    500 \times 64 \times 256
    =
    8{,}192{,}000
\end{equation}

next-token target positions. This fixed-update design matches the number of
optimizer steps and token-level training targets across the three
tokenizer-specific language models.

Validation performance is evaluated every 50 updates, and the checkpoint with
the lowest validation BPC is retained for evaluation. For all three methods,
the lowest observed validation BPC occurs at the final evaluated checkpoint,
update 500.

\subsection{Evaluation Metric}
\label{subsec:bpc}

Perplexity depends on the tokenizer because changing the segmentation changes
the units over which next-token probabilities are defined. It is therefore
reported only as a secondary, tokenizer-dependent diagnostic. The primary
evaluation metric is bits per character (BPC), which normalizes negative
log-likelihood by the number of characters in the original text.

For a held-out character sequence of length $C$ represented by token targets
$y_1,\ldots,y_M$, BPC is defined as

\begin{equation}
    \mathrm{BPC}
    =
    \frac{
        -\displaystyle\sum_{i=1}^{M}
        \log_2 p(y_i \mid y_{<i})
    }{
        C
    }.
    \label{eq:bpc}
\end{equation}

Because the denominator is the number of original characters rather than the
number of resulting tokens, BPC provides a common character-normalized scale
for comparing predictive performance across tokenizers with different
segmentations.

\section{Results}
\label{sec:results}

\subsection{Compute-Matched Language-Model Performance}
\label{subsec:fixed-update-results}

Table~\ref{tab:fixed500} presents the main fixed-update language-model
results. The three models use the same architecture, parameter count, batch
size, context length, optimizer settings, random seed, and number of optimizer
updates.

\begin{table}[H]
\centering
\caption{Fixed-update TinyGPT results after 500 optimizer updates per
language model. Lower BPC is better. Perplexity is reported as a secondary,
tokenizer-dependent metric.}
\label{tab:fixed500}
\begin{tabular}{lrrrr}
\toprule
Method &
Val. BPC &
Test BPC &
Val. PPL &
Test PPL \\
\midrule
Frequency BPE &
3.340716 &
3.349258 &
437.70 &
465.14 \\

SG-BPE &
3.299369 &
3.289352 &
323.31 &
351.73 \\

MDL-SG &
\textbf{3.261212} &
\textbf{3.243573} &
283.82 &
296.41 \\
\bottomrule
\end{tabular}
\end{table}

Under this fixed-update comparison, MDL-SG obtains the lowest observed BPC on
both validation and test data. Relative to frequency BPE, validation BPC
decreases from 3.340716 to 3.261212, corresponding to a relative reduction of

\begin{equation}
    \frac{
        3.340716-3.261212
    }{
        3.340716
    }
    \times 100
    =
    2.38\%.
\end{equation}

On the test set, BPC decreases from 3.349258 to 3.243573, corresponding to a
relative reduction of

\begin{equation}
    \frac{
        3.349258-3.243573
    }{
        3.349258
    }
    \times 100
    =
    3.16\%.
\end{equation}

The comparison with SG-BPE is especially relevant because SG-BPE supplies the
statistical ranking mechanism extended by MDL-SG. Relative to SG-BPE, the
observed validation BPC reduction is

\begin{equation}
    \frac{
        3.299369-3.261212
    }{
        3.299369
    }
    \times 100
    =
    1.16\%,
\end{equation}

and the corresponding test BPC reduction is

\begin{equation}
    \frac{
        3.289352-3.243573
    }{
        3.289352
    }
    \times 100
    =
    1.39\%.
\end{equation}

These differences are observed without increasing model size or the number of
optimizer updates. Each language model contains exactly 2,024,448 trainable
parameters and is trained for exactly 500 optimizer updates.

\subsection{Compression and Predictive Utility}
\label{subsec:compression-prediction}

Taken together, Tables~\ref{tab:tpc} and~\ref{tab:fixed500} show that stronger
sequence compression does not necessarily correspond to lower
character-normalized language-model loss. Frequency BPE achieves the lowest
test TPC,

\begin{equation}
    \mathrm{TPC}_{\mathrm{Freq}}
    =
    0.377960,
\end{equation}

whereas MDL-SG has

\begin{equation}
    \mathrm{TPC}_{\mathrm{MDL}}
    =
    0.395010.
\end{equation}

Thus, frequency BPE represents the same held-out text using fewer tokens.
Nevertheless, its test BPC is higher:

\begin{equation}
    3.349258
    \quad \text{versus} \quad
    3.243573.
\end{equation}

At this matched operating point, the tokenizer producing the strongest
sequence compression is therefore not the tokenizer producing the lowest
character-normalized predictive loss.

This result illustrates the distinction underlying the proposed method.
Frequency BPE directly rewards repeated compression opportunities, whereas
SG-BPE favors pairs with high Significance-Gain scores. MDL-SG adds two
further requirements: a candidate must pass the replication criterion on the
disjoint replication partition, and introducing the resulting token must
produce positive held-out MDL gain. The resulting vocabulary is less
compression-oriented at this operating point but yields lower observed
language-model BPC under the fixed training budget.

\subsection{Learning Curves Under the Fixed-Update Budget}
\label{subsec:learning-curves}

The observed ordering of the three methods is not limited to the final
checkpoint. Validation BPC decreases throughout the 500-update training run.
At update 200, the validation BPC values are 3.9821 for frequency BPE,
3.7657 for SG-BPE, and 3.7144 for MDL-SG. At update 300, the corresponding
values are 3.5712, 3.4661, and 3.4141. By update 500, they reach 3.3407,
3.2994, and 3.2612, respectively.

Over these later evaluated checkpoints, the observed ordering is

\begin{equation}
    \mathrm{BPC}_{\mathrm{MDL\text{-}SG}}
    <
    \mathrm{BPC}_{\mathrm{SG}}
    <
    \mathrm{BPC}_{\mathrm{Freq}}.
\end{equation}

Because the lowest validation BPC for every model occurs at the final
evaluated update, these experiments should be interpreted as a matched
500-update comparison rather than as evidence that the models have fully
converged. Moreover, the reported runs use a single common random seed, so
the observed differences should not be interpreted as estimates of
run-to-run statistical variability. Nevertheless, the fixed-update design
ensures that the observed advantage of MDL-SG is not attributable to
receiving more optimizer updates than the comparison models.

\section{Discussion}
\label{sec:discussion}

The experiments provide evidence for two distinct properties of the proposed
procedure. First, MDL-SG produces data-dependent stopping points rather than
using a predetermined merge count. Second, under the reported fixed-update,
single-seed language-model comparison, the vocabulary selected by MDL-SG
produces lower observed character-normalized loss than the matched
frequency-BPE and SG-BPE baselines.

\subsection{Why Statistical Replication Alone Is Not Sufficient}
\label{subsec:replication-not-sufficient}

The stopping-boundary analysis shows that MDL-SG does not terminate simply
because statistically supported candidate pairs disappear. At the
500K-character operating point, 6,564 discovery candidates remain after the
final accepted merge, and 1,890 of them pass the replication stage.
Nevertheless, every such candidate has non-positive held-out MDL gain.

This distinction is central to the proposed formulation. A candidate can pass
the one-sided replication test together with the per-iteration
Benjamini--Hochberg procedure at

\begin{equation}
    q=0.05
\end{equation}

while simultaneously satisfying

\begin{equation}
    G_{\mathrm{MDL}}(x,y) \leq 0.
\end{equation}

The first condition means that the candidate belongs to the rejection set
defined by the specified replication procedure. The second means that
converting the pair into an additional token does not reduce the held-out
description-length objective enough to offset its assigned representation
cost. Passing the statistical replication criterion and providing positive
held-out token utility are therefore related but distinct requirements.

The observed stopping points illustrate this distinction. MDL-SG terminates
after 209 merges for the 120K-character sample, 433 merges for the
250K-character sample, and 847 merges for the 500K-character sample. Across
these three tested corpus sizes, larger tokenizer-training samples are
associated with later stopping points, consistent with the interpretation that
more data can provide sufficient evidence to retain increasingly specialized
structures.

The largest rejected gain also moves closer to zero across the three tested
corpus sizes:

\begin{equation}
    -0.4223
    \;\longrightarrow\;
    -0.1507
    \;\longrightarrow\;
    -0.0474
    \quad \text{bits}.
\end{equation}

This pattern is consistent with a greedy procedure operating near the
positive-gain boundary. However, three corpus sizes are insufficient to
establish a general scaling law or monotonic relationship beyond the
experiments reported here.

\subsection{Why Lower TPC Does Not Imply Lower BPC}
\label{subsec:tpc-bpc-discussion}

The tokenizer-only and language-model results show that stronger sequence
compression does not necessarily coincide with lower character-normalized
predictive loss at the tested operating point.

Frequency BPE achieves the lowest test TPC,

\begin{equation}
    0.377960,
\end{equation}

compared with

\begin{equation}
    0.388895
    \quad \text{for SG-BPE}
\end{equation}

and

\begin{equation}
    0.395010
    \quad \text{for MDL-SG}.
\end{equation}

Relative to MDL-SG, frequency BPE therefore uses approximately

\begin{equation}
    \frac{
        0.395010-0.377960
    }{
        0.395010
    }
    \times 100
    \approx
    4.32\%
\end{equation}

fewer tokens per character on the test set.

Despite this compression advantage, frequency BPE has the highest observed
test BPC. MDL-SG reduces test BPC by 3.16\% relative to frequency BPE and by
1.39\% relative to SG-BPE in the reported fixed-update experiment. Thus,
sequence length alone does not determine character-normalized predictive
performance.

A frequency-based merge is directly rewarded for replacing many adjacent
occurrences and can therefore provide substantial sequence compression.
SG-BPE instead emphasizes unexpectedly cohesive pairs through the
Significance-Gain ranking. MDL-SG adds two further requirements: a candidate
must pass the replication criterion on the disjoint replication partition,
and its introduction must yield positive held-out MDL gain.

The resulting MDL-SG vocabulary preserves somewhat finer segmentation at the
matched vocabulary size. In the reported fixed-update experiment, this less
compression-oriented representation nevertheless yields lower observed BPC.
These results suggest that raw sequence compression and predictive usefulness
can favor different merge choices.

\subsection{Relationship to Significance-Gain BPE}
\label{subsec:relationship-sg}

MDL-SG is designed as an extension of SG-BPE rather than a replacement for
its statistical ranking principle \cite{nouri2026siggain}. The
Significance-Gain score continues to determine the order in which candidate
structures are considered. The principal change is that a high ranking is no
longer sufficient for acceptance.

The proposed procedure can be summarized as

\begin{equation}
\boxed{
\text{SG discovery}
\;\longrightarrow\;
\text{replication on a disjoint partition}
\;\longrightarrow\;
\text{held-out MDL utility}
}
\end{equation}

followed by the stopping rule

\begin{equation}
\boxed{
\text{stop if no candidate passing replication has }
G_{\mathrm{MDL}}>0.
}
\end{equation}

This decomposition addresses two limitations of relying on a
significance-based ranking alone. First, a candidate that appears unusually
cohesive in the discovery partition must also pass the specified replication
criterion on a separate, disjoint partition. Second, passing the replication
criterion is not sufficient for acceptance unless introducing the resulting
token also yields positive net held-out MDL gain.

The comparison with SG-BPE at the same 847-merge and 1,017-token
stored-vocabulary operating point provides the most direct empirical
comparison for assessing the additional replication and MDL stages. In the
reported fixed-update experiment, validation BPC decreases from 3.299369 for
SG-BPE to 3.261212 for MDL-SG, while test BPC decreases from 3.289352 to
3.243573. These correspond to observed relative reductions of 1.16\% and
1.39\%, respectively.

Because the comparison uses a single common random seed, these differences
should be interpreted as observed results under the reported experimental
configuration rather than as estimates of average performance across repeated
training runs.

\section{Limitations}
\label{sec:limitations}

The present experiments demonstrate the feasibility of the proposed stopping
criterion, but several limitations remain.

\paragraph{Single corpus.}
All reported experiments use WikiText-103. Evaluation on additional corpora,
domains, and languages is needed before drawing broader conclusions about the
generality of the selected stopping behavior or downstream performance.

\paragraph{Limited corpus-scale study.}
Automatic stopping is evaluated at 120K, 250K, and 500K tokenizer-training
characters. The observed stopping points increase from 209 to 433 to
847 merges across these three sample sizes. This pattern is consistent with
the interpretation that larger samples can support additional candidate
structures, but three operating points are insufficient to establish a
general scaling relationship.

\paragraph{Contiguous partitioning.}
The current implementation divides each tokenizer-training sample into
contiguous 70\%, 15\%, and 15\% discovery, replication, and utility regions.
Because the underlying corpus contains multiple articles and topics, these
contiguous regions may differ in their content distribution. Future work
should compare alternative deterministic partitioning strategies, such as
block-disjoint or interleaved splits, while ensuring that artificial adjacent
pairs are not created across block boundaries.

\paragraph{Single language-model seed.}
The reported TinyGPT comparison uses one common random seed, 42, across the
three tokenizer-specific language models. Using the same seed improves
comparability of the reported runs, but a single seed does not quantify
run-to-run variability. Repeated training with multiple independent seeds is
needed to estimate the variance and statistical stability of the observed BPC
differences.

\paragraph{Finite optimization budget.}
For all three models, the lowest observed validation BPC occurs at the final
evaluated checkpoint, update 500. The results therefore provide a controlled
comparison under a fixed optimization budget rather than evidence that the
models have fully converged. Longer matched-update experiments are needed to
determine whether the observed ordering persists later in training.

\paragraph{MDL modeling choices.}
The utility criterion combines an empirical sequence code length, a BIC-style
active-vocabulary penalty, and a fixed-width merge-rule cost. These choices
provide a transparent operational description-length objective, but they are
not the only possible MDL formulation. Alternative universal codes,
complexity penalties, or dictionary encodings could change the exact stopping
point and should be investigated through sensitivity analysis.

\paragraph{Greedy merge trajectory.}
MDL-SG retains the Significance-Gain ranking and accepts the first candidate
in that order that passes replication and has positive held-out MDL gain. It
does not globally maximize MDL gain over all candidates at an iteration or
over complete merge sequences. As in standard BPE, earlier greedy decisions
therefore influence which candidate structures are available at later
iterations.

\section{Conclusion}
\label{sec:conclusion}

This paper introduced MDL-Calibrated Significance-Gain Pair Encoding
(MDL-SG), a subword construction procedure in which statistical discovery,
replication on a disjoint corpus partition, and held-out utility evaluation
serve distinct roles. Significance-Gain determines the ranking of candidate
merges, the replication stage filters candidates that do not pass the
specified statistical criterion on separate data, and the MDL stage determines
whether introducing a replicated candidate yields sufficient held-out
description-length benefit to justify its added representation cost.

The resulting stopping rule does not require a predetermined target
vocabulary size or merge count. Instead, vocabulary growth terminates when
either no candidate passes the replication stage or candidates continue to
pass replication but none yields positive net held-out MDL gain. In the
reported WikiText-103 experiments, the same procedure stops after 209, 433,
and 847 merges for tokenizer-training samples of 120K, 250K, and 500K
characters, respectively.

At the 500K-character operating point, the automatically selected
847-merge MDL-SG vocabulary is compared with frequency BPE and SG-BPE at the
same merge count and stored-vocabulary size using the same TinyGPT
architecture. Under the reported fixed-update, single-seed experiment,
MDL-SG achieves validation/test BPC values of 3.2612/3.2436, compared with
3.2994/3.2894 for SG-BPE and 3.3407/3.3493 for frequency BPE.

The experiments also illustrate a distinction between sequence compression
and character-normalized predictive performance. Frequency BPE produces
shorter token sequences, whereas MDL-SG yields lower observed BPC under the
reported training configuration. These results suggest that the merge choices
that maximize raw sequence compression need not coincide with those that
produce the lowest downstream predictive loss.

Overall, the proposed framework provides a data-dependent mechanism for
controlling subword vocabulary growth by combining statistical candidate
ranking, replication on separate data, and an explicit held-out
description-length criterion. Evaluation on larger corpora, additional
domains and languages, multiple training seeds, and alternative MDL
formulations remains an important direction for future work.

\section*{Code Availability}

The source code used to implement MDL-Calibrated Significance-Gain Pair
Encoding, reproduce the tokenizer experiments, verify the automatic stopping
boundaries, and run the matched TinyGPT evaluations is publicly available in
the project repository:

\begin{center}
\href{https://github.com/Meetra21/LLM_24/blob/main/MDL-Calibrated\%20Significance-Gain\%20Pair\%20Encoding_Second\%20Paper.ipynb}
{\texttt{MDL-SG implementation and experiments}}
\end{center}
\cite{nouri2025ijca}
\cite{nouri2025sobel}
\cite{nouri2026stepcache}
\cite{nouri2025critical}


\end{document}